\documentclass[10pt, a4paper]{article}

\usepackage{lrec-coling2024} 
\usepackage{subcaption}

\title{Sequence-to-Sequence Language Models for Character and Emotion Detection in Dream Narratives}

 \name{Gustave Cortal$^{1, 2}$}

\address{         
$^1$Paris-Saclay University, CNRS, ENS Paris-Saclay, LMF, 91190, Gif-sur-Yvette, France\\
    $^2$Paris-Saclay University, CNRS, LISN, 91400, Orsay, France\\
          gustave.cortal@ens-paris-saclay.fr}

\abstract{
 The study of dreams has been central to understanding human (un)consciousness, cognition, and culture for centuries. Analyzing dreams quantitatively depends on labor-intensive, manual annotation of dream narratives. We automate this process through a natural language sequence-to-sequence generation framework. This paper presents the first study on character and emotion detection in the English portion of the open DreamBank corpus of dream narratives. Our results show that language models can effectively address this complex task. To get insight into prediction performance, we evaluate the impact of model size, prediction order of characters, and the consideration of proper names and character traits. We compare our approach with a large
language model using in-context learning. Our supervised models perform better while having 28 times fewer parameters. Our model and its generated annotations are made publicly available.
 \\ \newline \Keywords{character and emotion detection,
sequence-to-sequence language model,
quantitative analysis of dreams} }
\begin{document}

\maketitleabstract

\section{Introduction}

Dreams and their meanings have been the subject of extensive interest for centuries. Artemidorus (2nd century AD) laid the foundations for dream interpretation in his work \textit{Oneirocritica}, studying the content of dreams and proposing techniques for interpreting them \citep{oneiro}. In the 19th century, Sigmund Freud marked a turning point in understanding dreams with his book “The Interpretation of Dreams” \citep{freud}. Freud attributed specific meanings to recurring characters, objects, and scenarios in dreams, emphasizing themes of sexuality and aggressiveness. According to Freud, dreams are expressions of desires repressed during the waking state and relieve tensions caused by these repressed desires, thereby maintaining good sleep and bodily health.

The idea that dreams function as emotional conflict resolution is echoed in contemporary theories \citep{emotionregulation1,emotionregulation2}, particularly those considering dreams as a nightly therapeutic mechanism that identifies dreamers' fears and offers new perspectives for resolving conflicts \citep{emotionregulation2}. Other contemporary theories focus on the dream's role in memory consolidation \citep{memoryconsolidation} and its function in selectively forgetting irrelevant information to facilitate future learning \citep{selectiveforgetting1,selectiveforgetting2}. Dreams have also been likened to simulators that train individuals to better react to new situations, particularly potential threats \citep{simulation1}.

\begin{figure}[!ht]
    \centering
    \includegraphics[scale=0.75]{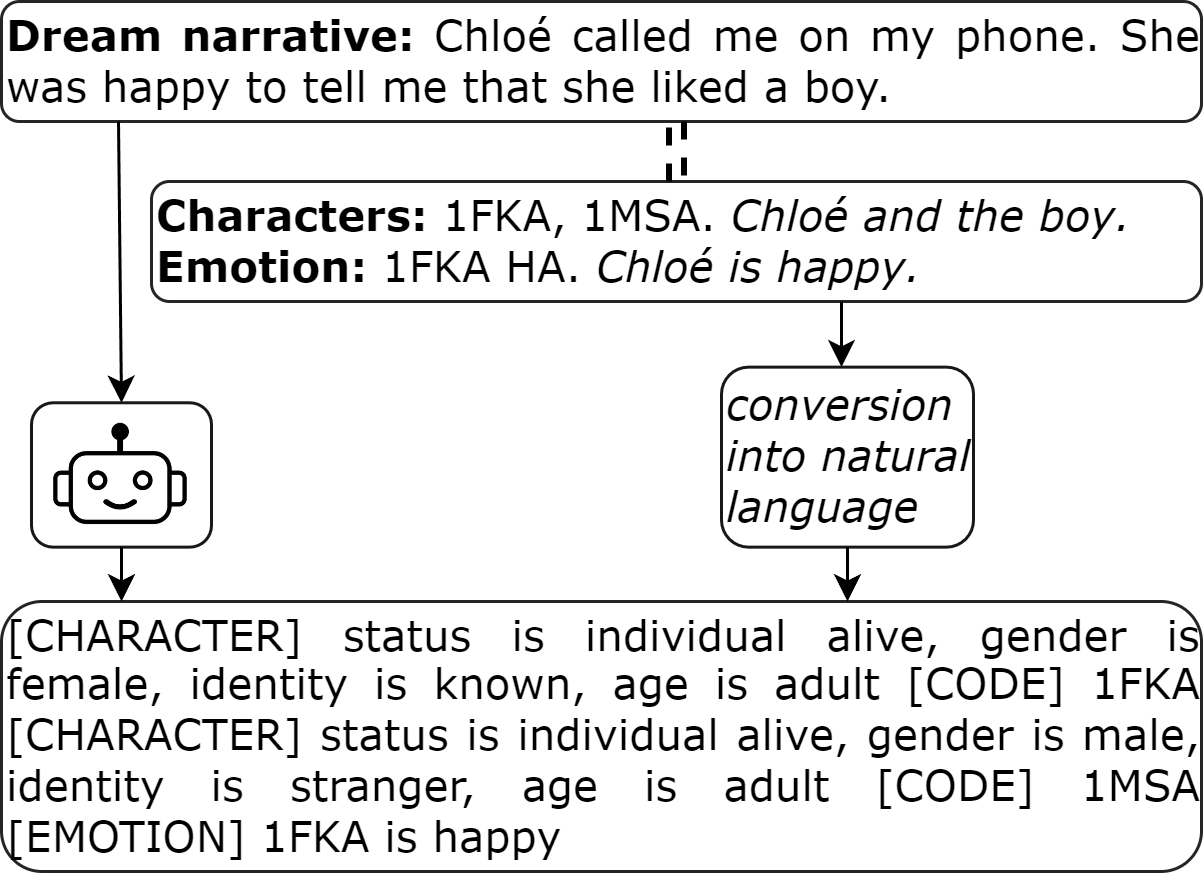}
    \caption{Sequence-to-sequence approach for automating the coding of dream narratives. Codes describing characters and their emotions are converted into natural language to produce the training data. From a narrative, a language model generates the natural language description of characters and their emotions.}
    \label{fig:approach}
\end{figure}


Limited empirical evidence supports the idea that dreaming contributes to adaptation and, consequently, the survival of individuals. However, recent studies offer evidence that dreams prolong events experienced during waking, known as the continuity hypothesis \citep{continuity}. The subfield of quantitative dream analysis has studied this hypothesis in detail, emphasizing the narratives' quantitative aspects by examining recurring patterns and associations between different narrative elements \citep{Zlotowicz73,scientificstudy}. Studies of this subfield collect and analyze objective data, like lexical fields \citep{Reinert93}, and do not rely on subjective or psychoanalytic interpretations. For example, if one wanted to study the theme of love in an individual's dreams, a straightforward approach would be to identify the frequency of terms related to the lexical field of love, such as “in love”, “romance”, and “passion”. If these terms appeared frequently, it could suggest that the individual focuses on the theme of love during the waking state. Quantitative dream analysis has thus brought to light elements related to daily interests and concerns in the waking state \citep{scientificstudy}.

Such an analytical approach requires a specific framework for systematically evaluating and comparing narratives. An annotation scheme clearly defining what to look for and how to note it ensures that all dreams will be analyzed according to the same criteria, allowing researchers to compare their results. Among the wide variety of annotation schemes for quantitative dream analysis \citep{dreamscale}, the Hall and Van de Castle (HVdC) scheme is the most commonly used \citep{hvdc,McNamara_Duffy-Deno_Marsh_Marsh_2019,ourdreams,Zheng2021,bertolini2023automatic}. This scheme identifies characters, emotions, interactions, and objects present in the narratives. However, applying this scheme to narratives is time-consuming due to its complexity and the need to train annotators for \emph{manual} dream analysis. Thus, although many available narratives exist, only a small subset has been annotated according to this scheme \citep{dreambank}. Automating the coding of narratives is an important challenge, as it would accelerate dream research by making thousands of annotated narratives available. In this paper, we propose to use transformer-based language models \citep{NIPS2017_3f5ee243} to automate the annotation of both characters and their emotions (\textit{i.e.}, not interactions and objects) according to the HVdC scheme.

There are few tools for automating the coding of dream narratives \citep{ourdreams,bertolini2023automatic}. \citet{bertolini2023automatic} are the first to use language models to predict the absence or presence of emotions, considering the entire narrative. However, the study does not account for the number of times an emotion appears in the narrative. Moreover, emotions are not identified concerning the characters experiencing them. 
 
We address these limitations by automatically identifying characters and their emotions. According to the HVdC scheme, characters are coded by symbols that classify their status, gender, identity concerning the dreamer, and age. For example, in the sentence “Emma is reading a book on the philosophy of language”, the character symbols for Emma are “1FKA”, with “1” for “individual alive”, “F” for “female”, “K” for “known”, and “A” for “adult”. We convert the symbols coding the characters and emotions into natural language so that language models can use their semantics in a text generation task. Using LaMini-Flan-T5, a sequence-to-sequence language model \citep{2020t5}, we generate a natural language description of characters and their emotions from a dream narrative, corresponding to the coded representation present in the HVdC scheme (see Figure \ref{fig:approach}). Our contributions are:

\begin{itemize}
    \item The joint prediction of characters and their emotions in dream narratives using a sequence-to-sequence language model. The description of characters and their emotions is leveraged by converting the corresponding codes into natural language;
    \item The examination of various phenomena, allowing insight into prediction performance, such as the effect of language model size, the order of character prediction, the conversion of codes into natural language, and the consideration of proper names and character traits;
    \item The comparison of our approach with a large language model using in-context learning. Our supervised models perform better while having 28 times fewer parameters;
    \item The release of our model\footnote{\url{https://huggingface.co/gustavecortal/dream-t5}} and the English part of the DreamBank corpus, including 27,952 annotated dream narratives.\footnote{\url{https://huggingface.co/datasets/gustavecortal/DreamBank-annotated}} Our model has accelerated the slow manual annotation of dream narratives. Researchers in quantitative analysis of dreams are encouraged to use our model to predict characters and emotions in unseen dreams. 
\end{itemize}

\section{Theoretical background and related works}

\subsection{Relationship with structured emotion and sentiment analysis\label{absa}}

Automated coding of dream narratives is closely related to the structured analysis of emotions, a task studied in NLP (Natural Language Processing). Inspired by the semantic role labeling task \citep{gildea-jurafsky-2000-automatic}, it aims to answer the question: “Who feels what, towards whom, and why?”. It identifies emotional cues with the entities feeling the emotions and the causes and targets of the emotions. \citet{campagnano-etal-2022-srl4e} propose a unified annotation scheme for different corpora of semantic roles related to emotions. Our task also shares similarities with aspect-based sentiment analysis, which seeks to identify the aspects of a product or subject and to determine the sentiment expressed about each of these aspects \citep{beneath}. For example, in the sentence “The battery life of this phone is incredible, but its camera quality is disappointing”, the sentiment is positive for the “battery life” aspect and negative for the “camera quality” aspect.

Dreams are often organized into a sequence of events that form a narrative. This narrative structure provides an exciting angle for analysis, and comparisons can be made with other types of narratives used in NLP, such as first-person emotional narratives \citep{cortal-etal-2023-emotion}. We believe that dream narratives can represent an interesting resource for emotion regulation assisted by NLP \citep{cortal2022natural}.

\subsection{Existing research on automated analysis of dream narratives}

\citet{clockssleep3030035} have summarized various works in NLP for the automated analysis of dream narratives. They can be divided into two approaches: dictionary and lexical database-based approaches \citep{Bulkeley2018,ourdreams,Mallett2021,Zheng2021,Yu2022}, and distributional semantic-based approaches \citep{ALTSZYLER2017178,10.3389/fnins.2018.00007,McNamara_Duffy-Deno_Marsh_Marsh_2019,GUTMANMUSIC2022103428}. We here present another approach that is based on recent language models \citep{bertolini2023automatic}.

\subsubsection{Dictionary and lexical database-based approach}

The dictionary-based approach analyzes the narrative word by word, referring to dictionaries and lexical databases. These dictionaries classify words into semantic categories. For example, the words “angry” and “frustration” can be associated with the “anger” category. Thus, for the sentence “I thought he was frustrated.”, the method will identify that the word “frustrated” belongs to the “anger” category. The emotion linked to this sentence will be anger. \citet{ourdreams} use the lexical database WordNet \citep{miller-1994-wordnet} to identify some aspects of the Hall and Van de Castle (HVdC) scheme, such as friendly and aggressive interactions.

\subsubsection{Language model-based approach}

The distributional semantic-based approach uses models for the vector representation of words or phrases. \citet{GUTMANMUSIC2022103428} identify prototypical situations of flight or attack through embedding and clustering narratives in a vector space with a pre-trained Sentence-BERT model \citep{reimers-2019-sentence-bert}. Sentence-BERT is a modification of BERT \citep{DBLP:conf/naacl/DevlinCLT19} that uses siamese networks for sentence embedding. 

Previous studies present two major limitations: a partial consideration of the context of the narratives and the absence of comparison with established coding systems like the HVdC scheme. To remedy these limitations, some studies combine the dictionary-based or distributional semantic-based approaches with machine learning \citep{McNamara_Duffy-Deno_Marsh_Marsh_2019,Yu2022}. For example, \citet{Yu2022} combines a sentiment dictionary with a support vector machine \citep{cortes1995support} to predict the overall sentiment of a dream narrative. \citet{bertolini2023automatic} are the first to finetune a pre-trained language model for detecting the absence or presence of emotions in narratives. This approach makes it possible to consider the whole context of the narrative and evaluate performance by comparing the model's predictions with HVdC gold-standard annotations. However, this study does not account for the frequency with which an emotion appears in the narrative. Moreover, emotions are not identified with the characters experiencing them. 

Generally, the studies we have cited focus mainly on the presence of emotional states in narratives without linking characters to these states. To remedy this limitation, we propose automatically identifying characters and their emotions using sequence-to-sequence language models based on the transformer architecture.

\section{Methodology}

In this section, we introduce the DreamBank dream narrative database~\cite{dreambank} and how characters and their emotions are coded according to the Hall and Van de Castle (HVdC) annotation scheme. We present our approach for training language models to generate characters and their emotions based on a dream narrative.

\subsection{DreamBank dream narrative database}


\begin{table}[ht]
\begin{center}
\begin{tabular}{l|l|l|l}
series &  information & years & nb \\
\hline
\textit{ed}         &  adult man &    1980-2002 & 143 \\
\textit{bea1}      &  teenager girl & 2003-2005  & 136 \\
\textit{b-baseline} & adult woman & 1960-1997  &  234 \\
\textit{emma}      &  adult woman & 1949-1997  &  285 \\
\textit{norms-m}   &  adult men & 1940s-1950s  & 485 \\
\textit{norms-f}   &  adult women & 1940s-1950s  &  483 \\
\end{tabular}
\end{center}
    \caption{Series of dreamers with the number of annotated narratives.}\label{series}
\end{table}

The open DreamBank database consists of 27,952 dream narratives in English and German.\footnote{A detailed description of DreamBank is available at \url{https://www.dreambank.net/grid.cgi}.} In this paper, we focus only on the English narratives, which are predominant and have been collected in the United States. Some of these narratives come from groups of individuals, such as university or college students, while others are a series of narratives from a single individual. Only 1,823 narratives have been annotated according to the HVdC scheme. We use a subset of 1,766 narratives with fewer than eight characters per narrative for training and evaluation.\footnote{We do not consider narratives with more than eight characters to avoid slowdown during the training and evaluation of our models, as these narratives tend to be very large.} Table \ref{series} presents the series of dreamers used.

\begin{table}[ht]  
\centering
   \begin{tabular}{|p{7cm}|}
   \hline
   \textbf{Narrative:} It was my birthday and I was having a party but in a place I've never been before. It was in a forest type area. All I remember is that at the same time I had two boyfriends. Only one was at my party, though he had just broken up with my best friend so I kinda felt uncomfortable being with him. We had got in an argument so he left. I don't quite remember how but we did make up but I don't remember when or why even got in the argument. I woke up when I heard the telephone ringing. 
   \\
   \hline
   \textbf{Coding:} 2MSC, 1MSC, 1FSC (\textit{two boyfriends, one boyfriend, my best friend})\\ 
   AP, D (\textit{dreamer has apprehension})\\
   \hline
    \end{tabular}
    \caption{A narrative and its coding. See the following section for the meaning of the symbols.}
    \label{tab:samplemma}
\end{table}

Table \ref{tab:samplemma} shows that in DreamBank, annotations are not anchored in texts. This makes our task akin to detecting implicit sentiments, a challenging task in sentiment analysis \cite{beneath}. This observation motivated our choice of a sequence-to-sequence generation-based approach. This approach takes a text as input (a dream narrative). It generates a new text as output (the natural language describing characters and their emotions) without requiring annotation anchors in texts. Our approach is illustrated in Figure \ref{fig:approach} and is explained in Section \ref{our_approach}.

\subsection{Hall and Van de Castle annotation scheme}

The HVdC scheme is one of the most popular schemes for analyzing and categorizing the content of dreams \cite{hvdc}. It classifies various elements of dream narratives, such as characters, emotions, interactions, objects, and locations.\footnote{Annotation guidelines are available at \url{https://dreams.ucsc.edu/Coding/}.} We use 1,766 narratives annotated according to this scheme and available from the DreamBank database. In this paper, we only consider the coding of characters and their emotions. Thus, we omit other annotations like interactions between characters or unlucky and lucky events for some characters. Considering other available annotations will be the subject of a future study.

\subsubsection{Character\label{character}}


In the HVdC scheme, characters can be people, animals, or creatures. For clarity and space, we focus only on people representing the majority in our annotated narratives (201 animals, 24 creatures, and 4,588 people), although our approach and models also consider animals and creatures. The majority of dreams have at least one character. Out of 1,766 annotated dreams, only 45 contain no characters. On average, there are 2.8 characters per narrative. Apart from animals and creatures, characters are coded according to four classes: status, gender, identity relative to the dreamer, and age. Each character is characterized by four symbols corresponding to the respective four classes. 

The “status” class indicates whether a character is an individual or a group of individuals. In addition, the status indicates whether a character is alive, dead, or imaginary. It also considers metamorphoses by considering the original and altered forms. The “gender” class has four subclasses: male, female, groups (with two genders), and unknown (gender not known by the dreamer or not clearly identified in the narrative). The “identity” class has eight subclasses arranged in a hierarchical order, from the most familiar to the least familiar: the dreamer's immediate family (\textit{e.g.}, parents or sister), the dreamer's relatives by marriage, blood, or adoption (\textit{e.g.}, cousin or aunt), characters directly known by the dreamer (\textit{e.g.}, roommates or boyfriend), characters known to the dreamer by their reputation (\textit{e.g.}, Winston Churchill or God), characters designated by their occupation (\textit{e.g.}, a student or a soldier), characters designated by their nationality, region, or city (\textit{e.g.}, a Frenchman), characters whose identity is not known by the dreamer (\textit{e.g.}, a girl or a crowd), and characters of which it is unknown if the dreamer knows their identity. Finally, the “age” class has four subclasses organized in a chronological decreasing order: adult, adolescent, child, and baby.

We observed that annotators sometimes rely on information that does not come from the narrative but rather from the characteristics of the dreamer, such as age and social status. For some narratives, it is impossible to determine the characters' ages from the text. However, it is possible to guess it by partially knowing the dreamer. For example, the characters in Béatrice's dreams (\textit{bea1} in Table \ref{series}), who is a teenager with several school friends in her dreams, are predominantly annotated as adolescents, even when no element in the text indicates it (for an example, see Table \ref{tab:samplemma}). We have decided to merge specific subclasses for age and identity to limit this bias. For the “age” class, the “baby” and “adolescent” subclasses are integrated into the “child” subclass. The “adult” subclass remains intact. For the “identity” class, the subclasses corresponding to the immediate family and relatives of the dreamer are integrated into the subclass of characters directly known by the dreamer, “known”. The subclass corresponding to characters of which it is unknown if the dreamer knows their identities is integrated into the subclass of unknown characters of the dreamer, “stranger”. The distribution of dreams according to the various subclasses for status, gender, identity, and age is illustrated in Figure \ref{fig:distrib_personnage} in the appendix. We summarize the subclasses and their corresponding symbols:

\begin{itemize}
    \item Status: individual alive (1), group alive (2), dead individual (3), dead group (4), imaginary individual (5), imaginary group (6), original form (7), changed form (8).
    \item Gender: male (M), female (F), joint (J), indefinite (I).
    \item Identity: known (K), prominent (P), occupational (O), ethnic (E), stranger (S).
    \item Age: adult (A), child (C).
\end{itemize}

\subsubsection{Emotion}

The HVdC scheme considers five emotional states: anger (AN), apprehension (AP), sadness (SD), confusion (CO), and happiness (HA). The coding procedure focuses mostly on explicit emotions.\footnote{The coding procedure is available at \url{https://dreams.ucsc.edu/Coding/emotions.html}.} It also indicates which characters are experiencing these emotions. For example, in the sentence “Emma is angry”, Emma's anger is coded by the symbols “1FKA AN”, denoting that an adult woman known by the dreamer (1FKA) is angry (AN). The dreamer is always present in the narrative and is represented by the symbol “D” when their emotion is coded. Our models are trained to predict characters and their corresponding emotions. We aim not only to predict that anger is present in the narrative but also to predict who is experiencing this anger. From the sentence “Emma is angry”, the language model should generate “1FKA AN”. In the following section, we will show that the generation of these symbols can be facilitated by converting them into natural language. This conversion allows for better exploitation of the semantics of references to characters and their emotions.

\begin{figure}[!htb]
    \centering
    \includegraphics[scale=0.4]{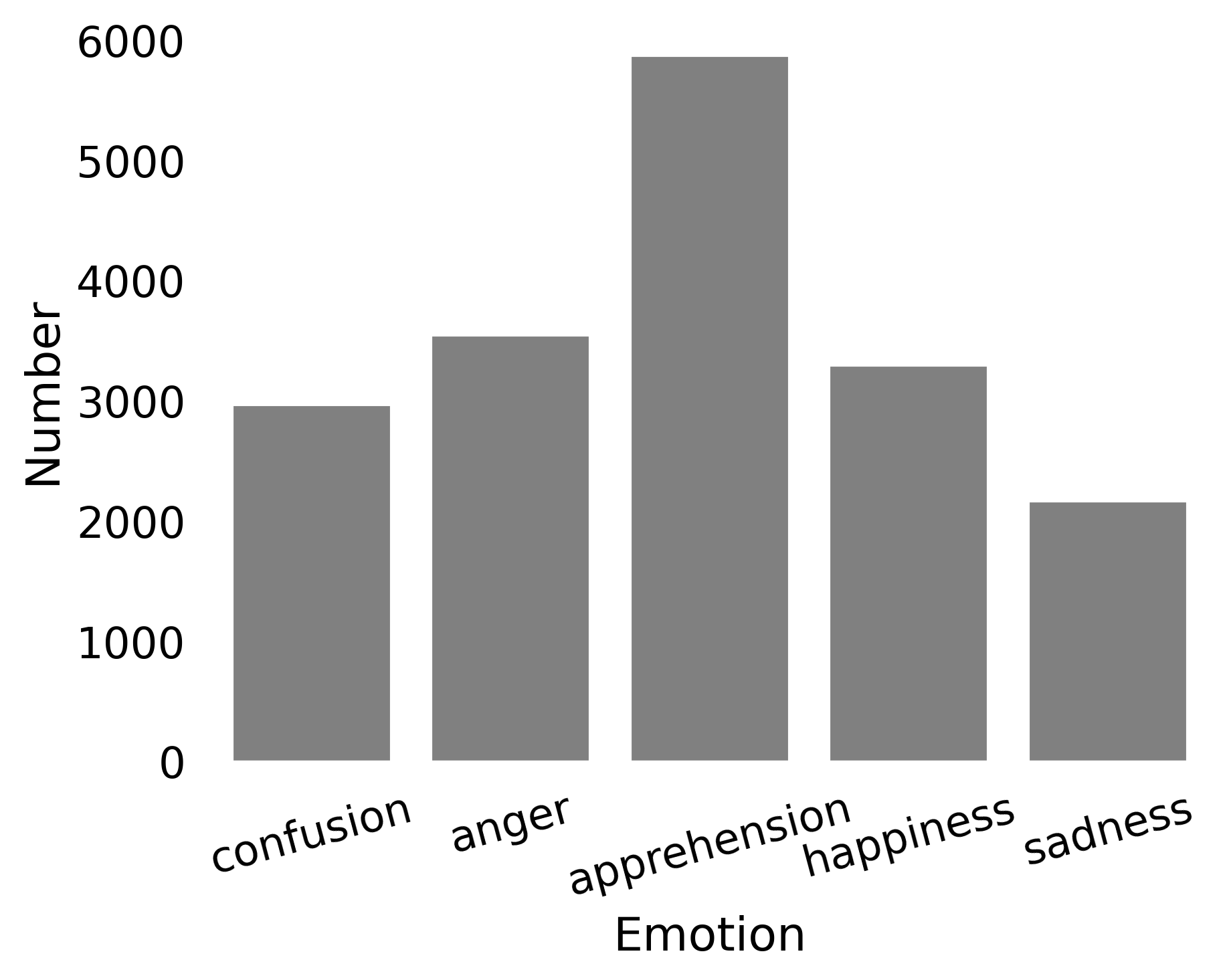}
    \caption{Distribution of emotional states.}
    \label{fig:emotional_states}
\end{figure}

Out of 1,766 narratives, 885 have no emotional content. Thus, emotionally neutral dreams are common. On average, narratives with emotional content have 1.6 emotions. Figure \ref{fig:emotional_states} describes the distribution of emotional states in the narratives. Apprehension is the dominant emotion. The dreamer experiences more emotions than other characters, with three-quarters of the emotions being attributed to the dreamer. 

\subsection{Description of our approach\label{our_approach}}

\subsubsection{Natural language conversion of character and emotion codes}

From a dream narrative, we generate the codes of characters and their emotions converted into natural language. Our approach, illustrated in Figure \ref{fig:approach}, is inspired by the study of \citet{zhang-etal-2021-aspect-sentiment} on aspect-based sentiment analysis. They detect all sentiment elements jointly by casting the task to a sequence-to-sequence generation process. Thanks to this approach, the task is solved end-to-end, and the semantics of the sentiment elements can be fully exploited by learning to generate them in the natural language form.

In DreamBank, narratives are annotated according to the HVdC scheme, which codes characters and their emotions with symbols. We propose to convert these symbols into natural language to exploit better the semantics of their references using language models. We jointly identify characters and their emotions and want to leverage the knowledge of pre-trained language models. The HVdC annotation guidelines map each symbol to its corresponding linguistic label. For example, for the character symbols “1FKA”, “1” becomes “individual alive”, “F” becomes “female”, “K” becomes “known”, and “A” becomes “adult”. The conversion of character symbols is summarized in Section \ref{character}. A similar conversion is also performed for emotions (\textit{e.g.}, “CO” becomes “confused”). We perform this conversion on all the symbols in our narratives. In this paper, we hypothesize that considering the semantics of references to characters and their emotions will improve prediction performance. Indeed, we believe a sequence-to-sequence language model can better exploit context to predict “individual alive female known adult” rather than “1FKA”.

As illustrated in Figure \ref{fig:approach}, to distinguish references to multiple characters, the marker “[CHARACTER]” is introduced. We apply the same strategy to separate emotions by introducing the marker “[EMOTION]”. To link the different subclasses, we introduce linguistic markers such as “status is”, “gender is”, “identity is”, and “age is”. We hypothesize that these linguistic markers will help our models correctly identify character classes. If a narrative does not contain a character (resp. does not contain an emotion), then the model must generate the sentence “There is no character.” (resp. “There is no emotion.”).

\subsubsection{Training}

We use an encoder-decoder language model based on the transformer architecture \cite{NIPS2017_3f5ee243} for training. We finetune LaMini-Flan-T5 \cite{lamini-lm}, a pre-trained T5 language model \cite{2020t5}. The model has 248 million parameters and was initially finetuned on 2.58 million instructions. The model takes as input a sequence of tokens and generates a new sequence of tokens as output. When writing this article, LaMini-Flan-T5 is one of the best language models with less than one billion parameters and has been evaluated on fifteen different NLP tasks \cite{lamini-lm}. We chose a small model size to ensure that the dream quantitative analysis community can easily reuse our supervised models to analyze novel dreams. We use a batch size of sixteen with a learning rate of $3^{-4}$. The number of epochs is set to fifteen for all experiments.\footnote{The full set of hyperparameters is made available at \url{https://huggingface.co/gustavecortal/dream-t5}.} Greedy decoding is used for inference.

The character and emotion generation procedure consists of two steps. The first step identifies the characters in the narrative with their status, gender, identity, and age. For each character, the model generates the corresponding natural language representation and then its code (\textit{e.g.}, “[CHARACTER] status is individual alive, gender is female, identity is known, age is adult [SYMBOL] 1FKA”). The second step relies on the first and associates certain identified characters with emotional states. The first step is essential; if the model does not correctly identify the characters, it will not correctly identify the emotional states related to them. In the example of Figure \ref{fig:approach}, the characters “1FKA” and “1MSA” (resp. Chloe and the boy) are identified during the first step. Then, in the second step, happiness is associated with the character “1FKA” since Chloe is happy. Thanks to our sequence-to-sequence approach, the model performs both steps end-to-end.

\subsubsection{Evaluation\label{evaluation}}

During the evaluation, the generated sequence corresponding to characters and their emotions in the natural language form is converted back to HVdC codes. For example, in the generated sequence “[CHARACTER] status is individual alive, gender is female, identity is known, age is adult [SYMBOL] 1FKA”, we extract the status “individual alive”, the gender “female”, the identity “known”, and the age “adult”. Then, we use the conversion presented in \ref{character} to find the character's corresponding symbols, namely “1”, “F”, “K”, and “A”. We proceed in the same way to recover emotion codes (\textit{e.g.}, from “[EMOTION] 1FKA is happy”, we get “1FKA HA”). The prediction is considered null if decoding fails because the generated sequence does not conform to the predefined format. The codes generated by our models are compared with the gold-standard codes by calculating recall, precision, and \( F1 \)-score. We only display the \( F1 \)-score to save space. 

The prediction of a character is considered correct if and only if its status, gender, identity, and age match the gold-standard symbols. We also display scores for predicting status, gender, identity, and age. These scores help us to refine the performance evaluation. The prediction of an emotion is considered correct if and only if the character code and the emotion code match the gold-standard codes. 

Evaluation is performed across all series of dreamers. To have a reliable evaluation, we need to evaluate our models on dreamers who are not from the training set. The models are trained while leaving out a series of dreamers each time for evaluation. Therefore, since we have six series of dreamers (as described in Table \ref{series}), we train six models for each experiment and average the evaluation results. This evaluation method prevents leakage of series' specificities. Indeed, it is possible that specific characteristics of dreamers in the training set (\textit{e.g.}, the way of expressing emotions or describing characters) may be found in the evaluation set, which would bias the evaluation. 

\section{Experiments}

\begin{table*}[ht]
    \centering
\begin{tabular}{l|cccccc}
model & status & gender & identity & age & character & emotion \\
\hline
\textsc{baseline} & 82.87 & 78.02 & 76.17 & 86.21 & 64.74 &  75.13 \\
\hline
\textsc{no\textsubscript{semantics}} & 71.37 & 56.54\textbf{*} & 61.0  & 90.51  & 41.79\textbf{*} &  75.79 \\
\textsc{no\textsubscript{names}} & 80.66\textbf{*}  & 74.32\textbf{**}  & 74.2  & 83.95\textbf{*}  & 60.93\textbf{**} &  73.04\textbf{*} \\
\hline
\textsc{size\textsubscript{small}} & 78.35\textbf{**}  & 72.13\textbf{**}  & 70.25\textbf{**}  & 81.66\textbf{**}  & 56.79\textbf{**} &  70.15\textbf{**} \\
\textsc{size\textsubscript{large}} & 84.51\textbf{*} & 80.3\textbf{**}  & 78.63\textbf{**}  & 87.29  & 67.63\textbf{**} & 74.71 \\
\hline
\textsc{first\textsubscript{group}} & 82.33  & 77.71  & 74.86  & 85.61  & 63.71 &  71.94 \\
\textsc{first\textsubscript{individual}} & 80.59\textbf{**}  & 76.14  & 74.22\textbf{*}  & 83.87\textbf{**} & 62.67 &  67.32 \\
\textsc{first\textsubscript{emotion}} & 83.92 & 78.74 & 77.06  & 87.63  & 64.97 &  72.03 \\
\hline
\textsc{conversion\textsubscript{comma}} & 84.02\textbf{**}  & 79.84\textbf{**}  & 77.67\textbf{**}  & 87.08\textbf{*}  & 66.69\textbf{**} &  73.68 \\
\textsc{conversion\textsubscript{marker}} & 82.39  & 78.45  & 76.53  & 86.09  & 65.44 &  74.36 \\
\hline
\textsc{StableBeluga\textsubscript{1}} &   43.95** &   39.76** &     31.25** &  56.16** &      15.65** & -\\
\textsc{StableBeluga\textsubscript{3}} &   52.44** &   46.49** &     38.46** &  63.88** &      21.06** & -\\
\textsc{StableBeluga\textsubscript{5}} &   55.89** &   46.29** &     42.61** &  63.73** &      24.86** & - \\
\hline
\textsc{cross-validation} & 86.28 & 81.9  & 79.51  & 89.52 & 68.64 &  76.18\\
\end{tabular}
    \caption{Results. \textbf{**}: p $<$ 0.05; \textbf{*}: p $<$ 0.1.}
    \label{tab:result}
\end{table*}

\subsection{Sequence-to-sequence language models}

Given that this task has not been previously explored, we construct a \textsc{baseline} model, which is a finetuned LaMini-Flan-T5 that uses the configuration we presented in the previous section. Figure \ref{fig:approach} illustrates a generation example. In our experiments, all models are trained with the same hyperparameters, starting from the LaMini-Flan-T5 model containing 248 million parameters. Only \textsc{small} and \textsc{large} have different parameters. To get insight into prediction performance, we investigate several phenomena, such as the effect of the language model size, character prediction order, the manner of converting codes into natural language, and the consideration of proper names and the semantics of character references. Here is a description of the model configurations:

\begin{itemize}

\item \textsc{no\textsubscript{semantics}}: We investigate the effect of the semantics of character references. \textsc{no\textsubscript{semantics}} does not consider the semantics of character references, as it directly predicts the symbols. Revisiting the example in Figure \ref{fig:approach}, the target text will be “[CHARACTER] 1FKA [CHARACTER] 1MSA [EMOTION] 1MSA is happy”.

\item \textsc{no\textsubscript{names}}: We study the effect of proper names in the narratives. We apply a named entity recognition model\footnote{\url{https://huggingface.co/Jean-Baptiste/roberta-large-ner-english}} to detect proper names, which are replaced by the specific token “[PER]”. For example, the sentence “Emma is angry at Robert.” would become “[PER1] is angry at [PER2].”.


\item \textsc{size}: We investigate the effect of language model sizes. \textsc{size\textsubscript{small}} and \textsc{size\textsubscript{large}} contain 77 and 783 million parameters, respectively.

\item \textsc{first}: We study the effect of different character prediction orders. \textsc{first\textsubscript{individual}} predicts individuals before groups, and \textsc{first\textsubscript{group}} predicts groups before individuals. \textsc{first\textsubscript{emotion}} also studies the effect of predicting emotions before predicting characters, which involves reversing the two steps in a generation.

\item \textsc{conversion}: We investigate the effect of different ways to convert character codes into natural language. \textsc{conversion\textsubscript{comma}} separates subclasses with commas (\textit{e.g.}, “individual alive, female, known, adult”). \textsc{conversion\textsubscript{marker}} separates subclasses with specific markers (\textit{e.g.}, “[STATUS] individual alive [GENDER] female [IDENTITY] known [AGE] adult”).

\item \textsc{cross-validation}: We perform a five-fold cross-series validation (80-20 split) to quantify to what extent models are likely to rely on the specificities of the training set series.

\end{itemize}

\subsection{Autoregressive language models}

To compare our approach, we also present an experiment with a large decoder-only language model without supervised training on DreamBank narratives. We focus on predicting characters, which, as we will see, is the most challenging task. We use StableBeluga\footnote{\url{https://huggingface.co/stabilityai/StableBeluga-7B}}, an autoregressive language model with seven billion parameters based on Llama 2 \cite{touvron2023llama}. StableBeluga has been adapted from Llama 2 on a dataset similar to Orca \cite{mukherjee2023orca} to mimic the reasoning process of GPT-4 \cite{openai2023gpt4}. As of November 2023, StableBeluga is one of the best open models available for research according to the Open LLM Leaderboard\footnote{\url{https://huggingface.co/spaces/HuggingFaceH4/open_llm_leaderboard}}. StableBeluga has 28 times more parameters than our T5 models. We use StableBeluga for inference with a custom prompt available in the appendix.

We perform in-context learning \cite{dong2022survey}. For a given narrative, we insert several randomly chosen narratives from other dreamers after the mention of "Assistant:". We experiment with one, three, and five examples. Our models are respectively \textsc{StableBeluga\textsubscript{1}}, \textsc{StableBeluga\textsubscript{3}}, and \textsc{StableBeluga\textsubscript{5}}.

\subsection{Results}


The results of our experiments are displayed in Table \ref{tab:result}.\ The \textit{character} metric evaluates the first step (\textit{i.e.}, characters prediction), and the \textit{emotion} metric evaluates the second step (\textit{i.e.}, predicting the emotional states of certain characters identified in the first step).\ As described in Section \ref{evaluation}, we display the average evaluation results across six series.\ There are differences between the series that we do not explain here to save space.\

Our \textsc{baseline} model performs well, with \( F1 \)-scores of 64.74 and 75.13 for \textit{character} and \textit{emotion}, respectively. Our results show that language models can effectively address the complex task of predicting characters and their associated emotions through our approach. In the following observations, we compare the results of our models to the \textsc{baseline} model. Except for \textsc{cross-validation}, we perform statistical significance testing using a Wilcoxon signed-rank test. 

\paragraph{How does \textsc{StableBeluga} perform compared to our supervised models?}

We focus on the \textit{character} metric. We observe that the number of in-context examples increases performance. Indeed, \textsc{StableBeluga\textsubscript{5}} scores 9.21 points higher compared to \textsc{StableBeluga\textsubscript{1}}. The best model with five examples performs poorly compared to \textsc{baseline} (-39.88 points). Analyzing several predictions, we observe that \textsc{StableBeluga} manages to follow the generation format but hallucinates certain subclasses (e.g., "identity is student" even though "student" is not a valid subclass). Moreover, the model is too sensitive to the in-context examples and tends to use mostly characters from the examples during generation. These phenomena are known limitations of in-context learning \cite{dong2022survey}. Thus, our supervised models perform better while having 28 times fewer parameters.

\paragraph{How does considering the semantic aspects of character references impact performance?} Considering the semantics of character references improves \textit{character} (\textsc{baseline} gains 22.95 points compared to \textsc{no\textsubscript{semantics}}).\ Therefore, converting the character codes into natural language allows the language model to leverage their semantics.\ Our approach benefits from the pre-training of language models that capitalize on context and encode some world knowledge.

\paragraph{How does the use of proper names impact performance?}

\textsc{no\textsubscript{names}} performs worse overall compared to \textsc{baseline} (resp. -3.81 points and -2.09 points for \textit{character} and \textit{emotion}).\ The most impacted subclass is gender (-3.7 points).\ Therefore, the model relies on proper names to predict characters and their emotions.

\paragraph{How do the model sizes impact performance?}

Scaling up the models improves \textit{character} (\textsc{size\textsubscript{large}} gains 10.84 points compared to \textsc{size\textsubscript{small}}).\ \textsc{size\textsubscript{large}} does not improve \textit{emotion} compared to \textsc{baseline}.\ Therefore, scaling up the models is an interesting direction for improving character prediction.

\paragraph{How does imposing an order in character prediction impact performance?}

We wondered whether predicting individuals before groups (\textsc{first\textsubscript{individual}}) or predicting groups before individuals (\textsc{first\textsubscript{group}}) could improve performance.\ We observe no statistical significance.\ Therefore, we can't conclude that a specific order is better.

\paragraph{How does the sequence of predicting emotions and characters impact performance?}

\textsc{first\textsubscript{emotion}} reverses the two steps by generating emotions before characters.\ We observe no statistical significance.\ We can't conclude that emotion prediction relies on character prediction.

\paragraph{How does converting codes into natural language impact performance?}

Converting character codes into natural language with commas increases character prediction performance (1.95 points) while converting character codes with markers has no statistical significance.\ Therefore, it is unnecessary to introduce linguistic markers such as “status is” and “gender is” to link the subclasses.\ These markers may induce spurious correlations with the input narratives, thereby reducing performance.

\paragraph{How do series-specific factors impact performance?}

By performing cross-series validation, we observe that overall performance increases.\ Hence, the model relies on the specificities of the training set series to predict new dreams.\ To have a reliable evaluation, we must evaluate our models on dreamers who do not come from the training set.\ That is what we have done for the other experiments.

\section{Discussions}

\subsection{Data bias}

The recounting of a dream is primarily influenced by memory, writing style, and the socio-economic category of the dreamer \cite{sociologiereve}. For example, the dreamer may add elements not initially present during the dream experience or only select the most important elements for themselves. Additionally, there are biases in the representation of dreamers, predominantly educated women in the United States. Finally, given the wide variety of linguistic markers that verbalize emotion, the HVdC scheme cannot capture the full emotional richness of the narratives, mainly focusing on explicit emotions. 

\subsection{Nature of dreams}

Dream narratives contain less sensory and conceptual information than waking state narratives. For example, in a waking state, one could describe several dimensions of a meal, such as its texture, taste, and smell. In a dream, these details may be vaguer or absent. Additionally, the laws of physics are often broken, \textit{e.g.}, characters may teleport or transform. Dream narratives also do not describe repeating events experienced in the waking state. They have their own structure. These phenomenological properties can challenge annotators and tools seeking to model them. Several debates in the dream research community relate to our study \cite{dreamstanford}. For example, do dreams really have emotional content? Can dreamers correctly identify characters? Can dreamers identify with characters other than themselves?

\subsection{Limitations and future directions}

We have only considered a subset of DreamBank, which consists of narratives annotated according to the HVdC scheme. Including non-annotated dreams, perhaps using unsupervised learning on the narratives, would be interesting.

We have not accounted for all available annotations in the HVdC scheme. Narratives are also annotated according to other classes, such as interactions between characters, objects in scenes, and lucky or unlucky events for certain characters. These annotations could improve the prediction performance of characters and their emotions. For instance, interactions could represent emotions' physiological and motor dimensions, while lucky or unlucky events may provoke emotions in certain characters. 

We impose an arbitrary order in predicting traits associated with character classes, while this task does not necessarily have any ordering properties. We conducted a preliminary experiment and found that the trait prediction order affects performance. The order set by the HVdC scheme (status, gender, identity, and age) does not necessarily maximize performance. What would be the performance if identity is predicted first and status last? We plan to study this phenomenon fully in a future study.

\section{Conclusion}

The quantitative analysis of dreams has relied almost exclusively on the manual and time-consuming annotation of dream narratives. This work offers a new approach that automates the coding of narratives by treating it as a natural language sequence-to-sequence generation task. This approach makes it possible to predict characters and their emotions simultaneously. Our results show that language models can effectively address this complex task. We investigate several phenomena, such as the effect of the language model size, the prediction order of characters, the method of converting codes into natural language, and the consideration of proper names and semantic references to characters. We compare our approach with a large language model using in-context learning. Our supervised models perform better while having 28 times fewer parameters. To accelerate dream research, we have made our model and the English part of the annotated DreamBank database available online. Researchers in quantitative analysis of dreams are encouraged to use our model to predict characters and emotions in unseen dreams.

\section{Ethics statement}

This research involves the automated analysis of dream narratives, a topic that could touch on personal issues for dreamers. All dream narratives used for this research are sourced from the publicly available DreamBank database and are anonymized to ensure privacy and confidentiality. The dreamers have given their consent to appear in the database.

\section{Acknowledgements}

This work was performed using HPC resources from GENCI-IDRIS (Grant 20XX-AD011014205).

\section{Bibliographical References}\label{sec:reference}

\bibliography{lrec-coling2024-example}

\begin{thebibliography}{45}
\expandafter\ifx\csname natexlab\endcsname\relax\def\natexlab#1{#1}\fi

\bibitem[{Altszyler et~al.(2017)Altszyler, Ribeiro, Sigman, and {Fernández
  Slezak}}]{ALTSZYLER2017178}
Edgar Altszyler, Sidarta Ribeiro, Mariano Sigman, and Diego {Fernández
  Slezak}. 2017.
\newblock \href {https://doi.org/https://doi.org/10.1016/j.concog.2017.09.004}
  {The interpretation of dream meaning: Resolving ambiguity using latent
  semantic analysis in a small corpus of text}.
\newblock \emph{Consciousness and Cognition}, 56:178--187.

\bibitem[{Bertolini et~al.(2023)Bertolini, Elce, Michalak, Bernardi, and
  Weeds}]{bertolini2023automatic}
Lorenzo Bertolini, Valentina Elce, Adriana Michalak, Giulio Bernardi, and Julie
  Weeds. 2023.
\newblock \href {http://arxiv.org/abs/2302.14828} {Automatic scoring of dream
  reports' emotional content with large language models}.

\bibitem[{Bulkeley and Graves(2018)}]{Bulkeley2018}
Kelly Bulkeley and Mark Graves. 2018.
\newblock \href {https://doi.org/10.1037/drm0000071} {Using the {LIWC} program
  to study dreams.}
\newblock \emph{Dreaming}, 28(1):43--58.

\bibitem[{Campagnano et~al.(2022)Campagnano, Conia, and
  Navigli}]{campagnano-etal-2022-srl4e}
Cesare Campagnano, Simone Conia, and Roberto Navigli. 2022.
\newblock \href {https://doi.org/10.18653/v1/2022.acl-long.314} {{SRL4E} {--}
  {S}emantic {R}ole {L}abeling for {E}motions: {A} unified evaluation
  framework}.
\newblock In \emph{Proceedings of the 60th Annual Meeting of the Association
  for Computational Linguistics (Volume 1: Long Papers)}, pages 4586--4601,
  Dublin, Ireland. Association for Computational Linguistics.

\bibitem[{Cartwright(2005)}]{emotionregulation1}
Rosalind Cartwright. 2005.
\newblock \href
  {https://doi.org/https://doi.org/10.1016/B0-72-160797-7/50052-5} {Chapter 45
  - dreaming as a mood regulation system}.
\newblock In Meir~H. Kryger, Thomas Roth, and William~C. Dement, editors,
  \emph{Principles and Practice of Sleep Medicine (Fourth Edition)}, fourth
  edition edition, pages 565--572. W.B. Saunders, Philadelphia.

\bibitem[{Cortal et~al.(2022)Cortal, Finkel, Paroubek, and
  Ye}]{cortal2022natural}
Gustave Cortal, Alain Finkel, Patrick Paroubek, and Lina Ye. 2022.
\newblock \href {https://inria.hal.science/hal-03805702} {{Natural Language
  Processing for Cognitive Analysis of Emotions}}.
\newblock In \emph{{Semantics, Memory, and Emotion 2022}}, Paris, France.

\bibitem[{Cortal et~al.(2023)Cortal, Finkel, Paroubek, and
  Ye}]{cortal-etal-2023-emotion}
Gustave Cortal, Alain Finkel, Patrick Paroubek, and Lina Ye. 2023.
\newblock \href {https://aclanthology.org/2023.latechclfl-1.8} {Emotion
  recognition based on psychological components in guided narratives for
  emotion regulation}.
\newblock In \emph{Proceedings of the 7th Joint SIGHUM Workshop on
  Computational Linguistics for Cultural Heritage, Social Sciences, Humanities
  and Literature}, pages 72--81, Dubrovnik, Croatia. Association for
  Computational Linguistics.

\bibitem[{Cortes and Vapnik(1995)}]{cortes1995support}
Corinna Cortes and Vladimir Vapnik. 1995.
\newblock Support-vector networks.
\newblock \emph{Machine learning}, 20(3):273--297.

\bibitem[{Crick and Mitchison(1983)}]{selectiveforgetting1}
Francis Crick and Graeme Mitchison. 1983.
\newblock \href {https://doi.org/10.1038/304111a0} {The function of dream
  sleep}.
\newblock \emph{Nature}, 304(5922):111--114.

\bibitem[{Crick and Mitchison(1995)}]{selectiveforgetting2}
Francis Crick and Graeme Mitchison. 1995.
\newblock \href {https://doi.org/10.1016/0166-4328(95)00006-f} {{REM} sleep and
  neural nets}.
\newblock \emph{Behavioural Brain Research}, 69(1-2):147--155.

\bibitem[{Devlin et~al.(2019)Devlin, Chang, Lee, and
  Toutanova}]{DBLP:conf/naacl/DevlinCLT19}
Jacob Devlin, Ming{-}Wei Chang, Kenton Lee, and Kristina Toutanova. 2019.
\newblock \href {https://doi.org/10.18653/V1/N19-1423} {{BERT:} pre-training of
  deep bidirectional transformers for language understanding}.
\newblock In \emph{Proceedings of the 2019 Conference of the North American
  Chapter of the Association for Computational Linguistics: Human Language
  Technologies, {NAACL-HLT} 2019, Minneapolis, MN, USA, June 2-7, 2019, Volume
  1 (Long and Short Papers)}, pages 4171--4186. Association for Computational
  Linguistics.

\bibitem[{Diekelmann and Born(2010)}]{memoryconsolidation}
Susanne Diekelmann and Jan Born. 2010.
\newblock \href {https://doi.org/10.1038/nrn2762} {The memory function of
  sleep}.
\newblock \emph{Nature Reviews Neuroscience}, 11(2):114--126.

\bibitem[{Domhoff(2003)}]{scientificstudy}
G.~William Domhoff. 2003.
\newblock \href {https://doi.org/10.1037/10463-000} {\emph{The scientific study
  of dreams: Neural networks, cognitive development, and content analysis.}}
\newblock American Psychological Association.

\bibitem[{Domhoff and Schneider(2008)}]{dreambank}
G.~William Domhoff and Adam Schneider. 2008.
\newblock \href {https://doi.org/10.1016/j.concog.2008.06.010} {Studying dream
  content using the archive and search engine on {DreamBank}.net}.
\newblock \emph{Consciousness and Cognition}, 17(4):1238--1247.

\bibitem[{Dong et~al.(2022)Dong, Li, Dai, Zheng, Wu, Chang, Sun, Xu, Li, and
  Sui}]{dong2022survey}
Qingxiu Dong, Lei Li, Damai Dai, Ce~Zheng, Zhiyong Wu, Baobao Chang, Xu~Sun,
  Jingjing Xu, Lei Li, and Zhifang Sui. 2022.
\newblock A survey on in-context learning.
\newblock \emph{arXiv preprint arXiv: 2301.00234}.

\bibitem[{Elce et~al.(2021)Elce, Handjaras, and Bernardi}]{clockssleep3030035}
Valentina Elce, Giacomo Handjaras, and Giulio Bernardi. 2021.
\newblock \href {https://doi.org/10.3390/clockssleep3030035} {The language of
  dreams: Application of linguistics-based approaches for the automated
  analysis of dream experiences}.
\newblock \emph{Clocks \& Sleep}, 3(3):495--514.

\bibitem[{Flanagan(1966)}]{hvdc}
H.~M. Flanagan. 1966.
\newblock \href {https://doi.org/10.1192/bjp.112.490.963} {The content analysis
  of dreams. by calvin s. hall and robert l. van de castle new york: The
  century psychology series. 1966. pp. 320. price not given.}
\newblock \emph{The British Journal of Psychiatry}, 112(490):963–964.

\bibitem[{Fogli et~al.(2020)Fogli, Aiello, and Quercia}]{ourdreams}
Alessandro Fogli, Luca~Maria Aiello, and Daniele Quercia. 2020.
\newblock \href {https://doi.org/10.1098/rsos.192080} {Our dreams, our selves:
  automatic analysis of dream reports}.
\newblock \emph{Royal Society Open Science}, 7(8):192080.

\bibitem[{Freud(1983)}]{freud}
Sigmund Freud. 1983.
\newblock The interpretation of dreams.
\newblock In \emph{Literature and Psychoanalysis}, pages 29--33. Columbia
  University Press.

\bibitem[{Gildea and Jurafsky(2000)}]{gildea-jurafsky-2000-automatic}
Daniel Gildea and Daniel Jurafsky. 2000.
\newblock \href {https://doi.org/10.3115/1075218.1075283} {Automatic labeling
  of semantic roles}.
\newblock In \emph{Proceedings of the 38th Annual Meeting of the Association
  for Computational Linguistics}, pages 512--520, Hong Kong. Association for
  Computational Linguistics.

\bibitem[{{Gutman Music} et~al.(2022){Gutman Music}, Holur, and
  Bulkeley}]{GUTMANMUSIC2022103428}
Maja {Gutman Music}, Pavan Holur, and Kelly Bulkeley. 2022.
\newblock \href {https://doi.org/https://doi.org/10.1016/j.concog.2022.103428}
  {Mapping dreams in a computational space: A phrase-level model for analyzing
  fight/flight and other typical situations in dream reports}.
\newblock \emph{Consciousness and Cognition}, 106:103428.

\bibitem[{Harris{-}McCoy(2012)}]{oneiro}
Daniel~E. Harris{-}McCoy. 2012.
\newblock \emph{Artemidorus' Oneirocritica: Text, Translation, and Commentary}.
\newblock Oxford University Press.

\bibitem[{Lahire(2021)}]{sociologiereve}
B.~Lahire. 2021.
\newblock \href {https://books.google.fr/books?id=xzQNEAAAQBAJ}
  {\emph{L'interpr{\'e}tation sociologique des r{\^e}ves}}.
\newblock Poche / Sciences humaines et sociales. La D{\'e}couverte.

\bibitem[{Mallett et~al.(2021)Mallett, Picard-Deland, Pigeon, Wary, Grewal,
  Blagrove, and Carr}]{Mallett2021}
Remington Mallett, Claudia Picard-Deland, Wilfred Pigeon, Madeline Wary, Alam
  Grewal, Mark Blagrove, and Michelle Carr. 2021.
\newblock \href {https://doi.org/10.1007/s42761-021-00080-8} {The relationship
  between dreams and subsequent morning mood using self-reports and text
  analysis}.
\newblock \emph{Affective Science}, 3(2):400--405.

\bibitem[{McNamara et~al.(2019)McNamara, Duffy-Deno, Marsh, and
  Marsh}]{McNamara_Duffy-Deno_Marsh_Marsh_2019}
Patrick McNamara, Kelly Duffy-Deno, Tom Marsh, and Thomas~Jr. Marsh. 2019.
\newblock \href {https://doi.org/10.11588/ijodr.2019.1.48744} {Dream content
  analysis using artificial intelligence}.
\newblock \emph{International Journal of Dream Research}, 12(1):42–52.

\bibitem[{Miller(1994)}]{miller-1994-wordnet}
George~A. Miller. 1994.
\newblock \href {https://aclanthology.org/H94-1111} {{W}ord{N}et: A lexical
  database for {E}nglish}.
\newblock In \emph{{H}uman {L}anguage {T}echnology: Proceedings of a Workshop
  held at {P}lainsboro, {N}ew {J}ersey, {M}arch 8-11, 1994}.

\bibitem[{Mukherjee et~al.(2023)Mukherjee, Mitra, Jawahar, Agarwal, Palangi,
  and Awadallah}]{mukherjee2023orca}
Subhabrata Mukherjee, Arindam Mitra, Ganesh Jawahar, Sahaj Agarwal, Hamid
  Palangi, and Ahmed Awadallah. 2023.
\newblock \href {http://arxiv.org/abs/2306.02707} {Orca: Progressive learning
  from complex explanation traces of gpt-4}.

\bibitem[{OpenAI(2023)}]{openai2023gpt4}
OpenAI. 2023.
\newblock Gpt-4 technical report.
\newblock \emph{PREPRINT}.

\bibitem[{Poria et~al.(2023)Poria, Hazarika, Majumder, and Mihalcea}]{beneath}
Soujanya Poria, Devamanyu Hazarika, Navonil Majumder, and Rada Mihalcea. 2023.
\newblock \href {https://doi.org/10.1109/TAFFC.2020.3038167} {Beneath the tip
  of the iceberg: Current challenges and new directions in sentiment analysis
  research}.
\newblock \emph{IEEE Transactions on Affective Computing}, 14(1):108--132.

\bibitem[{Raffel et~al.(2020)Raffel, Shazeer, Roberts, Lee, Narang, Matena,
  Zhou, Li, and Liu}]{2020t5}
Colin Raffel, Noam Shazeer, Adam Roberts, Katherine Lee, Sharan Narang, Michael
  Matena, Yanqi Zhou, Wei Li, and Peter~J. Liu. 2020.
\newblock \href {http://jmlr.org/papers/v21/20-074.html} {Exploring the limits
  of transfer learning with a unified text-to-text transformer}.
\newblock \emph{Journal of Machine Learning Research}, 21(140):1--67.

\bibitem[{Reimers and Gurevych(2019)}]{reimers-2019-sentence-bert}
Nils Reimers and Iryna Gurevych. 2019.
\newblock \href {http://arxiv.org/abs/1908.10084} {Sentence-bert: Sentence
  embeddings using siamese bert-networks}.
\newblock In \emph{Proceedings of the 2019 Conference on Empirical Methods in
  Natural Language Processing}. Association for Computational Linguistics.

\bibitem[{Reinert(1993)}]{Reinert93}
Max Reinert. 1993.
\newblock \href {https://www.persee.fr/doc/lsoc_0181-4095_1993_num_66_1_2632}
  {Les "mondes lexicaux" et leur "logique" à travers l'analyse statistique
  d'un corpus de récits de cauchemars}.
\newblock \emph{Langage \& société}, 66:5--39.

\bibitem[{Sanz et~al.(2018)Sanz, Zamberlan, Erowid, Erowid, and
  Tagliazucchi}]{10.3389/fnins.2018.00007}
Camila Sanz, Federico Zamberlan, Earth Erowid, Fire Erowid, and Enzo
  Tagliazucchi. 2018.
\newblock \href {https://doi.org/10.3389/fnins.2018.00007} {The experience
  elicited by hallucinogens presents the highest similarity to dreaming within
  a large database of psychoactive substance reports}.
\newblock \emph{Frontiers in Neuroscience}, 12.

\bibitem[{Schredl and Hofmann(2003)}]{continuity}
Michael Schredl and Friedrich Hofmann. 2003.
\newblock \href {https://doi.org/10.1016/s1053-8100(02)00072-7} {Continuity
  between waking activities and dream activities}.
\newblock \emph{Consciousness and Cognition}, 12(2):298--308.

\bibitem[{Thill and Svensson(2011)}]{simulation1}
Serge Thill and Henrik Svensson. 2011.
\newblock The inception of simulation: a hypothesis for the role of dreams in
  young children.
\newblock In \emph{Proceedings of the Annual Meeting of the Cognitive Science
  Society}, volume~33.

\bibitem[{Touvron et~al.(2023)Touvron, Martin, Stone, Albert, Almahairi,
  Babaei, Bashlykov, Batra, Bhargava, Bhosale, Bikel, Blecher, Ferrer, Chen,
  Cucurull, Esiobu, Fernandes, Fu, Fu, Fuller, Gao, Goswami, Goyal, Hartshorn,
  Hosseini, Hou, Inan, Kardas, Kerkez, Khabsa, Kloumann, Korenev, Koura,
  Lachaux, Lavril, Lee, Liskovich, Lu, Mao, Martinet, Mihaylov, Mishra,
  Molybog, Nie, Poulton, Reizenstein, Rungta, Saladi, Schelten, Silva, Smith,
  Subramanian, Tan, Tang, Taylor, Williams, Kuan, Xu, Yan, Zarov, Zhang, Fan,
  Kambadur, Narang, Rodriguez, Stojnic, Edunov, and Scialom}]{touvron2023llama}
Hugo Touvron, Louis Martin, Kevin Stone, Peter Albert, Amjad Almahairi, Yasmine
  Babaei, Nikolay Bashlykov, Soumya Batra, Prajjwal Bhargava, Shruti Bhosale,
  Dan Bikel, Lukas Blecher, Cristian~Canton Ferrer, Moya Chen, Guillem
  Cucurull, David Esiobu, Jude Fernandes, Jeremy Fu, Wenyin Fu, Brian Fuller,
  Cynthia Gao, Vedanuj Goswami, Naman Goyal, Anthony Hartshorn, Saghar
  Hosseini, Rui Hou, Hakan Inan, Marcin Kardas, Viktor Kerkez, Madian Khabsa,
  Isabel Kloumann, Artem Korenev, Punit~Singh Koura, Marie-Anne Lachaux,
  Thibaut Lavril, Jenya Lee, Diana Liskovich, Yinghai Lu, Yuning Mao, Xavier
  Martinet, Todor Mihaylov, Pushkar Mishra, Igor Molybog, Yixin Nie, Andrew
  Poulton, Jeremy Reizenstein, Rashi Rungta, Kalyan Saladi, Alan Schelten, Ruan
  Silva, Eric~Michael Smith, Ranjan Subramanian, Xiaoqing~Ellen Tan, Binh Tang,
  Ross Taylor, Adina Williams, Jian~Xiang Kuan, Puxin Xu, Zheng Yan, Iliyan
  Zarov, Yuchen Zhang, Angela Fan, Melanie Kambadur, Sharan Narang, Aurelien
  Rodriguez, Robert Stojnic, Sergey Edunov, and Thomas Scialom. 2023.
\newblock \href {http://arxiv.org/abs/2307.09288} {Llama 2: Open foundation and
  fine-tuned chat models}.

\bibitem[{Vaswani et~al.(2017)Vaswani, Shazeer, Parmar, Uszkoreit, Jones,
  Gomez, Kaiser, and Polosukhin}]{NIPS2017_3f5ee243}
Ashish Vaswani, Noam Shazeer, Niki Parmar, Jakob Uszkoreit, Llion Jones,
  Aidan~N Gomez, \L~ukasz Kaiser, and Illia Polosukhin. 2017.
\newblock \href
  {https://proceedings.neurips.cc/paper_files/paper/2017/file/3f5ee243547dee91fbd053c1c4a845aa-Paper.pdf}
  {Attention is all you need}.
\newblock In \emph{Advances in Neural Information Processing Systems},
  volume~30. Curran Associates, Inc.

\bibitem[{Walker and van~der Helm(2009)}]{emotionregulation2}
Matthew~P. Walker and Els van~der Helm. 2009.
\newblock \href {https://doi.org/10.1037/a0016570} {Overnight therapy? the role
  of sleep in emotional brain processing.}
\newblock \emph{Psychological Bulletin}, 135(5):731--748.

\bibitem[{Windt(2021)}]{dreamstanford}
Jennifer~M. Windt. 2021.
\newblock {Dreams and Dreaming}.
\newblock In Edward~N. Zalta, editor, \emph{The {Stanford} Encyclopedia of
  Philosophy}, {S}ummer 2021 edition. Metaphysics Research Lab, Stanford
  University.

\bibitem[{Winget and Kramer(1979)}]{dreamscale}
Carolyn Winget and Milton Kramer. 1979.
\newblock \emph{Dimensions of dreams}.
\newblock Gainesville: University of Florida Press.

\bibitem[{Wu et~al.(2023)Wu, Waheed, Zhang, Abdul-Mageed, and Aji}]{lamini-lm}
Minghao Wu, Abdul Waheed, Chiyu Zhang, Muhammad Abdul-Mageed, and Alham~Fikri
  Aji. 2023.
\newblock Lamini-lm: A diverse herd of distilled models from large-scale
  instructions.
\newblock \emph{CoRR}, abs/2304.14402.

\bibitem[{Yu(2022)}]{Yu2022}
Calvin Kai-Ching Yu. 2022.
\newblock \href {https://doi.org/10.1037/drm0000189} {Automated analysis of
  dream sentiment{\textemdash}the royal road to dream dynamics?}
\newblock \emph{Dreaming}, 32(1):33--51.

\bibitem[{Zhang et~al.(2021)Zhang, Deng, Li, Yuan, Bing, and
  Lam}]{zhang-etal-2021-aspect-sentiment}
Wenxuan Zhang, Yang Deng, Xin Li, Yifei Yuan, Lidong Bing, and Wai Lam. 2021.
\newblock \href {https://doi.org/10.18653/v1/2021.emnlp-main.726} {Aspect
  sentiment quad prediction as paraphrase generation}.
\newblock In \emph{Proceedings of the 2021 Conference on Empirical Methods in
  Natural Language Processing}, pages 9209--9219, Online and Punta Cana,
  Dominican Republic. Association for Computational Linguistics.

\bibitem[{Zheng and Schweickert(2021)}]{Zheng2021}
Xiaofang Zheng and Richard Schweickert. 2021.
\newblock \href {https://doi.org/10.1037/drm0000173} {Comparing hall van de
  castle coding and linguistic inquiry and word count using canonical
  correlation analysis.}
\newblock \emph{Dreaming}, 31(3):207--224.

\bibitem[{Zlotowicz(1973)}]{Zlotowicz73}
Michel Zlotowicz. 1973.
\newblock \href
  {https://www.persee.fr/doc/bupsy_0007-4403_1973_num_26_305_10419} {Sur
  l'analyse du cauchemar enfantin}.
\newblock \emph{Bulletin de psychologie}, 26(305):615--621.

\end{thebibliography}
\bibliographystyle{lrec-coling2024-natbib}




\section{Appendix}

\subsection{Prompt for \textsc{StableBeluga}}

\begin{table}[!htb]
    \centering
\footnotesize{
\begin{verbatim}
### System:
You are StableBeluga, an AI
that follows instructions extremely well.
Help as much as you can. You know the 
Hall and Van de Castle annotation scheme.
### User:
Classify CHARACTERS (status,
gender, identity, and age) in a DREAM
REPORT.\nGiven a DREAM REPORT, you must
follow the  format: CHARACTERS:
[CHARACTER]status is <status>, gender is 
<gender>, identity is  <identity>, age
is <age>\nWhere: <status> must be in
{"1":"individual alive", "2":"group
alive", "3":"dead individual", "4":
"dead group", "5":"imaginary individual",
"6":"imaginary group", "7": "original
form", "8":"changed form"}\n<gender> must
be in {"M":"male", "F":"female", "J":
"joint", "I":"indefinite"}\n<age> must be
in {"A":"adult", "C":"child"}\n<identity>
must be in {"K":"known", "P":"prominent",
"O":"occupational", "E":"ethnic", "S":
"stranger"}\nUse [CHARACTERS] to separate
multiple characters. Do not classify
the dreamer.
### Assistant:
\end{verbatim}
}
    \caption{\textsc{StableBeluga} prompt for character prediction.}
    \label{tab:prompt}
\end{table}

\normalsize

\subsection{Distribution of character classes}

\begin{figure}[!htb]
     \centering
     \begin{subfigure}[b]{0.4\textwidth}
         \centering
         \includegraphics[scale=0.4]{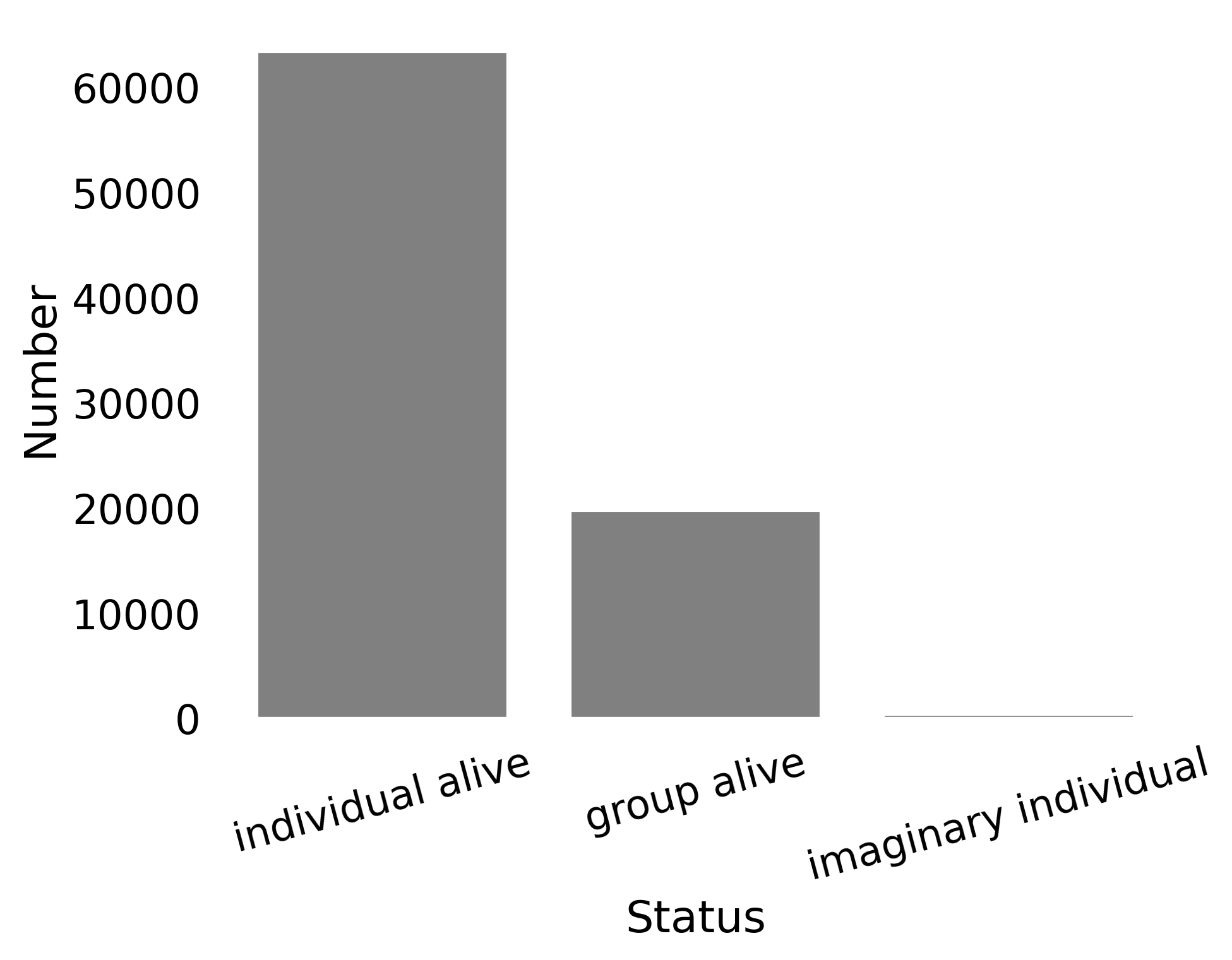}
         \caption{Status}
         \label{fig:status}
     \end{subfigure}
     \hspace{1cm}
     \begin{subfigure}[b]{0.4\textwidth}
         \centering
         \includegraphics[scale=0.4]{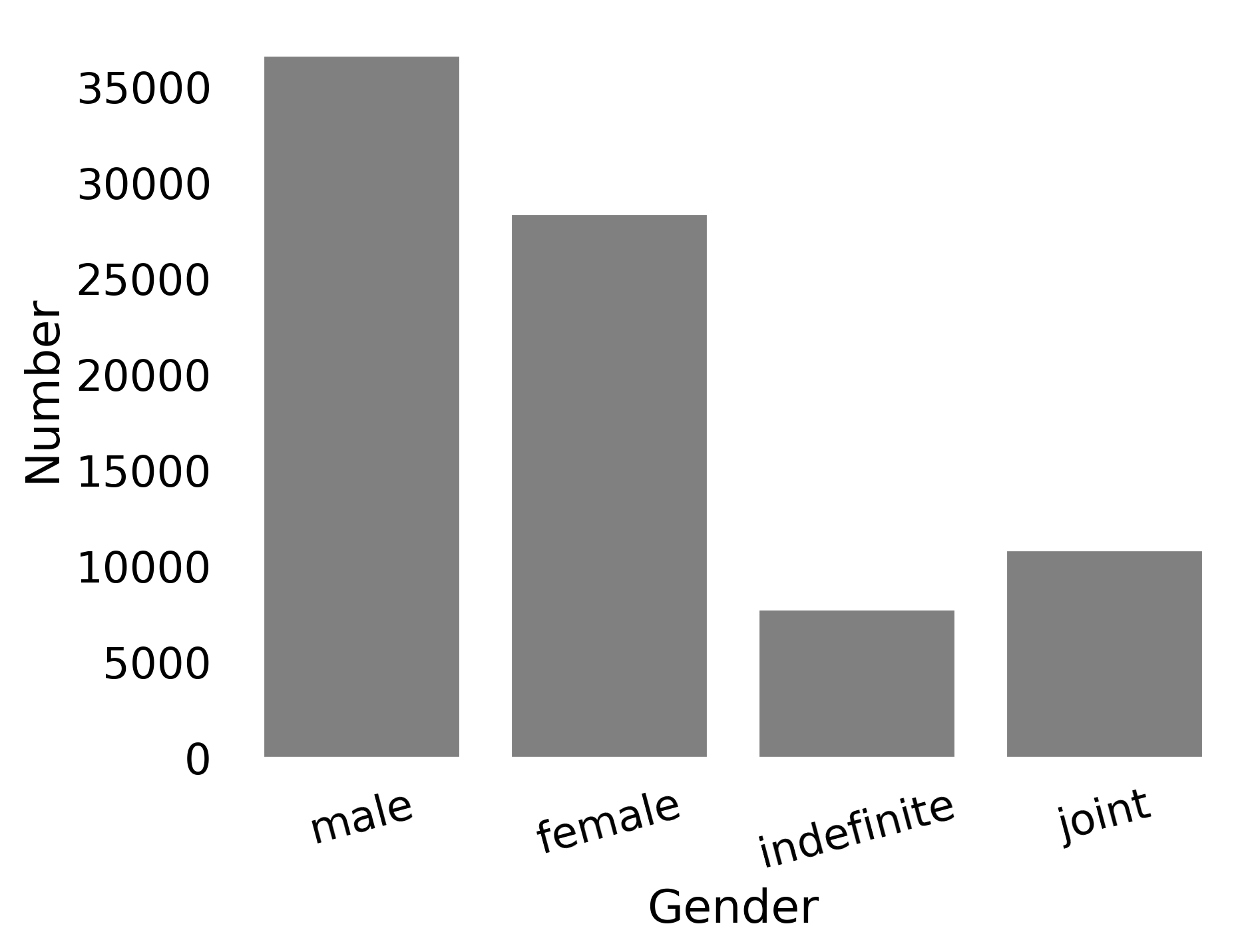}
         \caption{Gender}
         \label{fig:genre}
     \end{subfigure}
     \hfill
     \begin{subfigure}[b]{0.45\textwidth}
         \centering
         \includegraphics[scale=0.4]{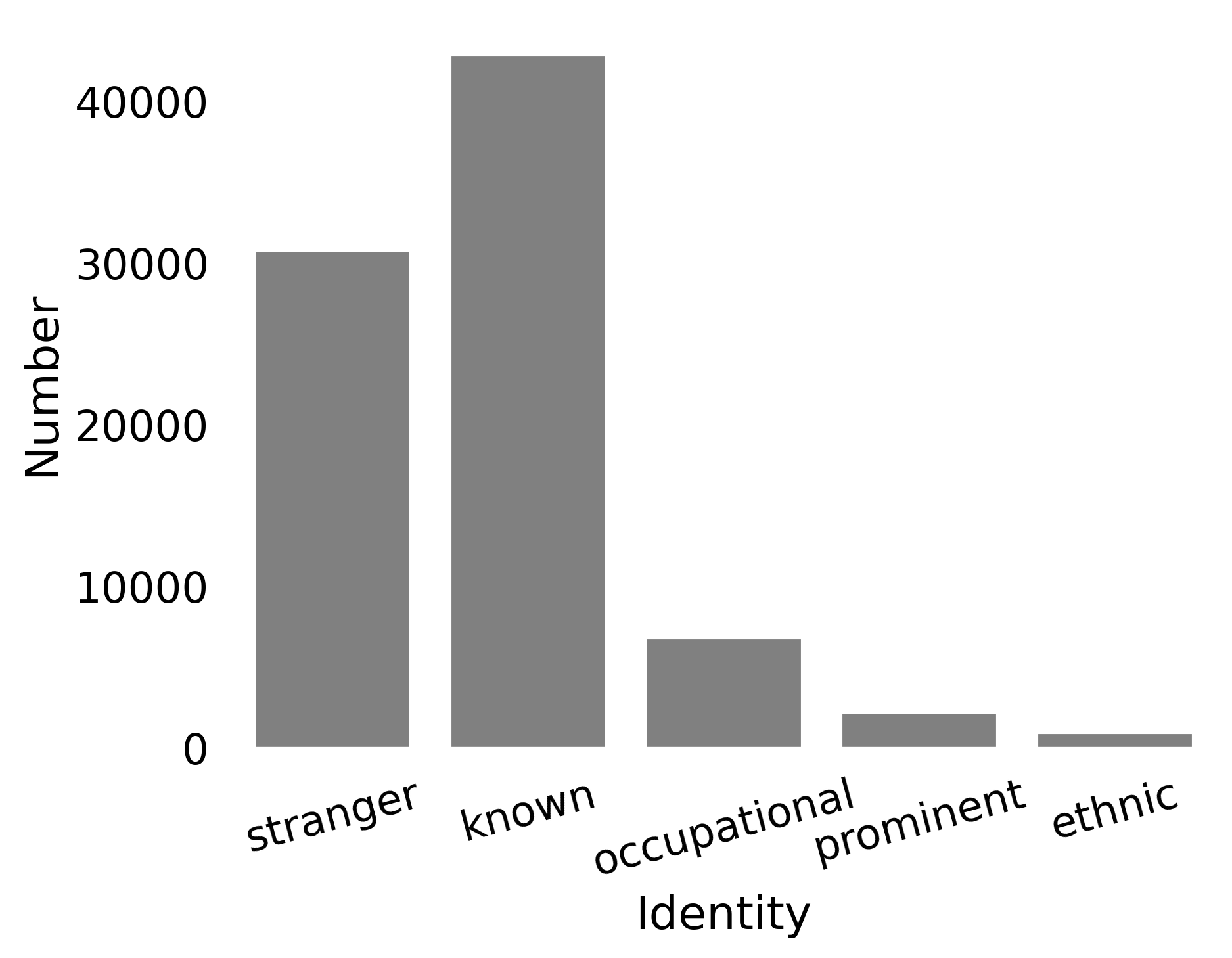}
         \caption{Identity}
         \label{fig:identite}
     \end{subfigure}
     \hspace{1cm}
          \begin{subfigure}[b]{0.4\textwidth}
         \centering
         \includegraphics[scale=0.4]{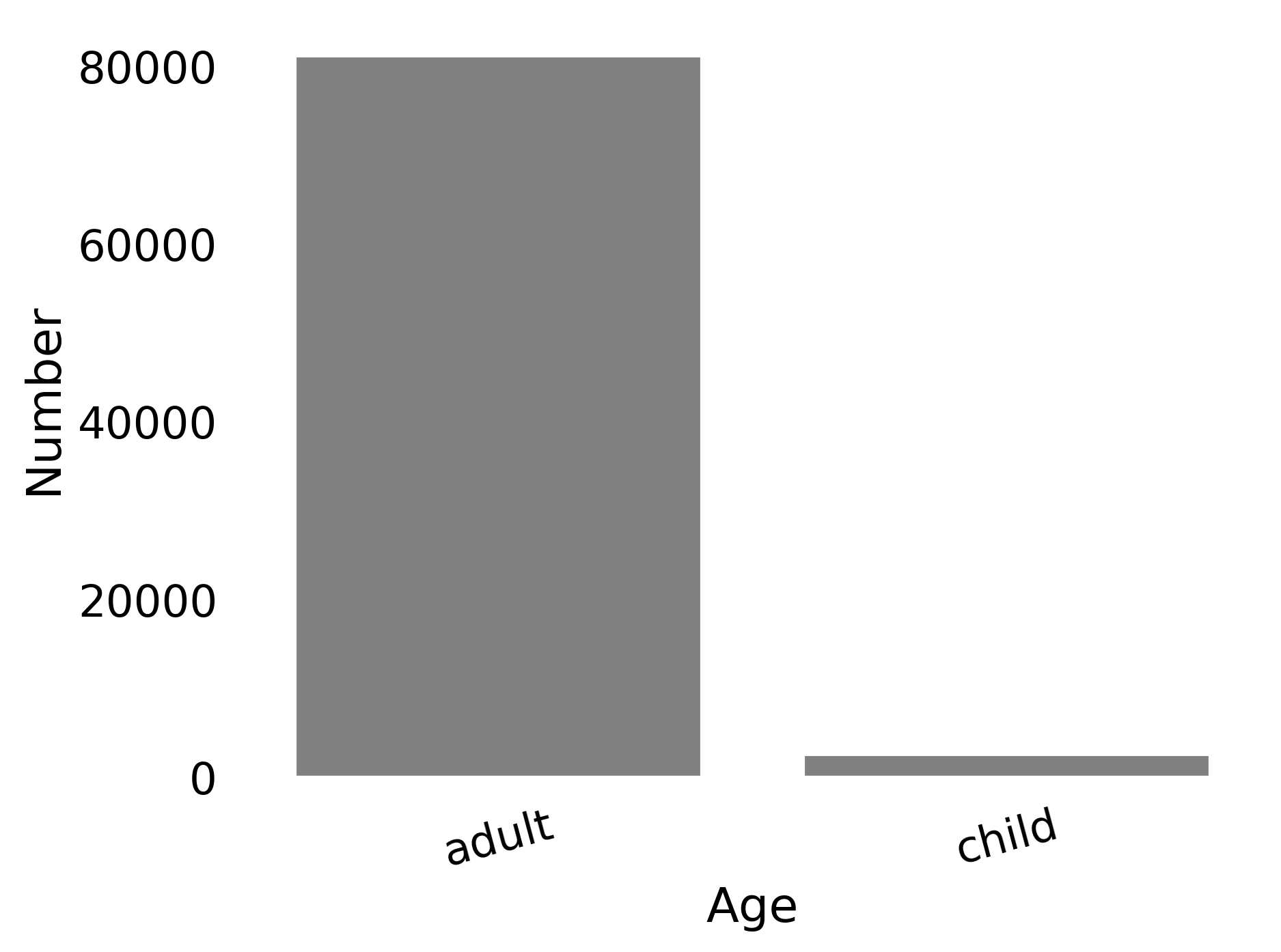}
         \caption{Age}
         \label{fig:age}
     \end{subfigure}
        \caption{Distribution of status \textbf{(a)}, gender \textbf{(b)}, identity \textbf{(c)} and age \textbf{(d)} in the dream narratives.} 
        \label{fig:distrib_personnage}
\end{figure}

\end{document}